\documentclass[10pt,twocolumn,letterpaper]{article}

\usepackage[dvipdfmx]{color}
\usepackage{comment}

\usepackage{cvpr}
\usepackage{times}
\usepackage{epsfig}
\usepackage{graphicx}
\usepackage{amsmath}
\usepackage{amssymb}

\usepackage[pagebackref=true,breaklinks=true,letterpaper=true,colorlinks,bookmarks=false]{hyperref}

\cvprfinalcopy 

\def\cvprPaperID{1960} 

\ifcvprfinal\fi

\newcommand{\En}{\underset{n}{\mathrm{E}\,}}
\newcommand{\VarPi}{\underset{P_i}{\mathrm{Var}}\,}
\newcommand{\CovPi}{\underset{P_i}{\mathrm{Cov}}\,}

\begin{document}

\title{Noise-Level Estimation from Single Color Image \\ Using Correlations Between Textures in RGB Channels}

\author{Akihiro Nakamura \\
The University of Tokyo \\
Tokyo, Japan \\
{\tt\small nakamura@mi.t.u-tokyo.ac.jp}
\and
Michihiro Kobayashi\\
Morpho, Inc.\\
Tokyo, Japan \\
{\tt\small m-kobayashi@morphoinc.com}
}

\maketitle

\begin{abstract}
We propose a simple method for estimating noise level from a single color image.
In most image-denoising algorithms, 
an accurate noise-level estimate results in good denoising performance;
however, it is difficult to estimate noise level from a single image 
because it is an ill-posed problem.
We tackle this problem by using prior knowledge that 
textures are highly correlated between RGB channels and noise is uncorrelated to other signals.
We also extended our method for RAW images
because they are available in almost all digital cameras and often used in practical situations.
Experiments show the high noise-estimation performance of our method
in synthetic noisy images.
We also applied our method to natural images including RAW images
and achieved better noise-estimation performance than conventional methods.
\end{abstract}

\section{Introduction}
Noise-level estimation is an important research area in computer vision
and has many applications such as image denoising \cite{NL-means, BM3D, baysian_denoising, nuclear_norm_minimization}
and edge detection \cite{edge_detection}.
It is important for these applications to obtain a good noise-level estimate in advance
because their performance strongly depends on the accuracy of this estimate.
However, noise-level estimation from a single image is fundamentally an ill-posed problem,
and it is impossible to separate noise from textures in a single image without using prior knowledge.
To tackle this problem, many methods have been developed, such as PCA-based and learning-based ones,
but most do not exploit the relationship between channels in color images.

We focus on the high correlations between channels in color images
and propose a noise-level estimation method using an assumption 
that textures are highly correlated between RGB channels 
and noise is uncorrelated with other signals (Figure \ref{tab:noise_correlations}).
We also extended our method for RAW images 
because they are often used in practical situations
and available in almost every digital camera.
We applied the proposed method to images artificially degraded with Gaussian noise
and succeeded in accurately estimating the noise level.
We also applied our method to natural noisy images including RAW images,
and it achieved better noise-estimation performance than conventional methods.
\begin{figure}[t]
    \centering
    \includegraphics[width=0.8\linewidth]{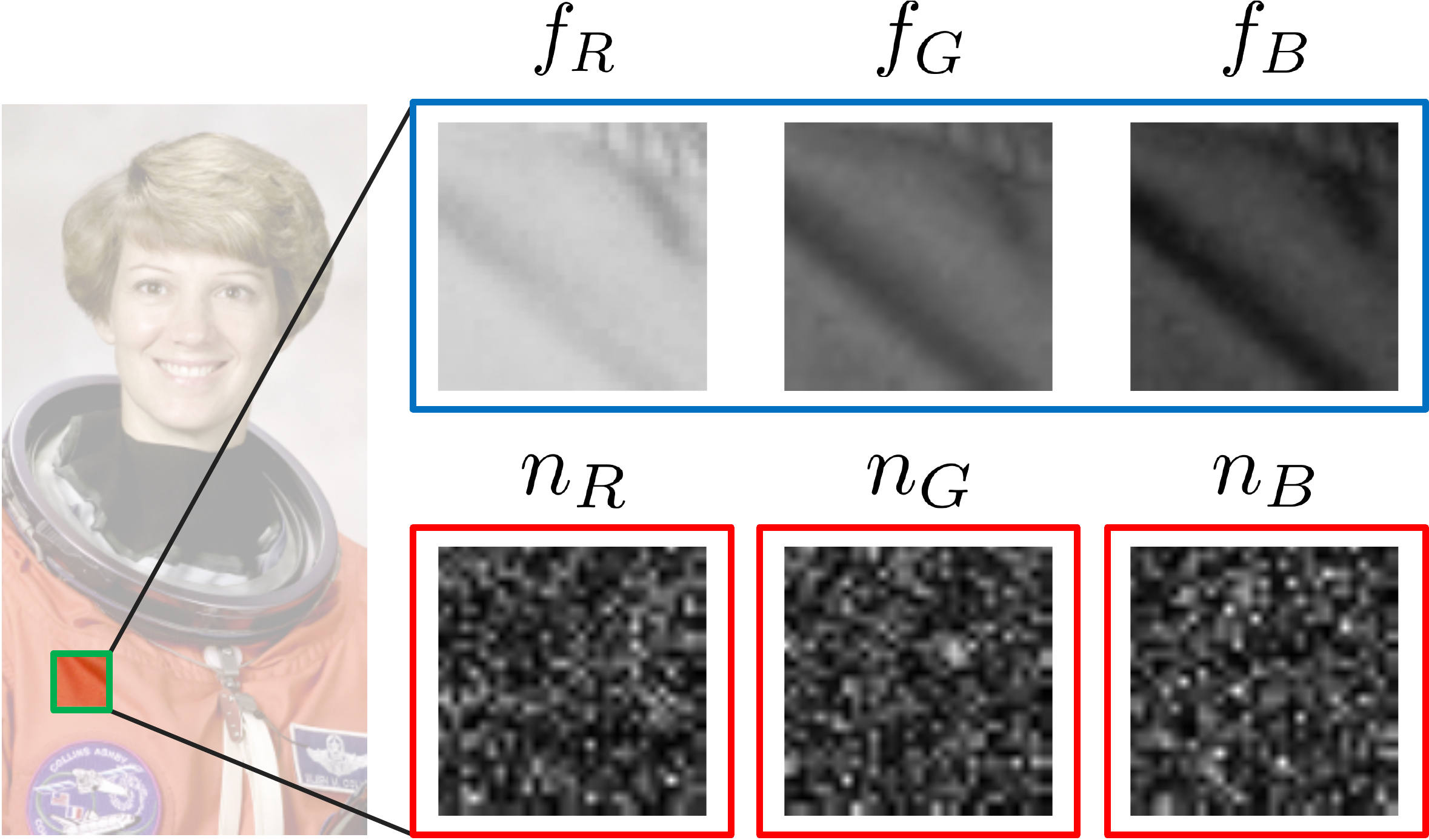}
    \caption{
        Left: With proposed method, small patches are sampled from input image.
        Right: Observed image is composed of noise-free image $f_c$ and noise $n_c \ (c\in \{R, G, B\})$.
        We assume that noise-free image $f_c$ is correlated between RGB channels
        and noise $n_c$ is uncorrelated with other signals.
    }
    \label{fig:texture_correlations}
\end{figure}

The main contribution of this paper is:
\begin{itemize}
	\item We propose a method of estimating noise level in a single color image using an assumption 
	that noise-free pixel values are often correlated between RGB channels while noise is uncorrelated.
	We believe we are the first to propose a {\it channel-correlation-based} method.
\end{itemize}

\section{Related Work}
Noise-level estimation from a single image is an ill-posed problem
because it is impossible to completely separate textures from noise,
and many methods have been developed to tackle this problem 
\cite{noise_estimate_1, noise_estimate_3, noise_estimate_2, noise_estimate_4}.
Some methods succeeded in estimating noise level 
by using patches sampled from homogeneous areas
\cite{homogeneous_detection, mean_deviation, adaptive_edge_detection, super_pixel}.
These methods are based on the assumption 
that there is a sufficient amount of flat areas in the input image,
but this assumption does not necessarily hold in natural images with rich textures.
Another method approximates noise 
by taking the difference between the original image and blurred image \cite{diff_smoothed}.
However, the noise level is often overestimated with these methods
because high-frequency components of textures still remain in the difference image.
PCA-based methods have been proposed \cite{PCA_1, PCA_2} to avoid these problems.
The core idea of PCA-based methods is that textures lie in a low-dimensional subspace
and the noise level can be estimated using eigenvalues of the redundant dimensions.
However, Chen \etal \cite{improved_PCA} pointed out that the noise level is underestimated with these methods.
They further investigated this problem and 
succeeded in improving the performance of PCA-based methods
by statistically analyzing the eigenvalues of the redundant space \cite{improved_PCA}.
The PCA-based methods discussed above work well when the image is degraded with white noise
but fail to estimate noise level accurately if the image contains non-white noise.
This is because non-white noise does not distribute uniformly in the redundant dimensions.

A fast patch-based noise-estimation method has recently been proposed \cite{weakly_textured}
using the Canny edge detector \cite{canny} 
to exclude highly textured areas.
This method is fast because of its simplicity, 
but the parameters of the edge detector have to be properly set
to correctly detect areas with rich textures.
Learning-based noise-estimation methods \cite{CNN_estimation} and
denoising methods \cite{CNN_denoising, DnCNN} 
using convolutional neural networks \cite{alex_net} have also been proposed recently.
These methods achieve high performance in noise estimation and denoising 
but suffer from high computational costs of convolutions for real-time computing 
when sufficient computational resources are not available such as in smartphones.

We propose a noise-level-estimation method
using an assumption that textures are highly correlated between RGB channels while noise is uncorrelated with other signals.
The proposed method does not require
flat areas in the input image,
the assumption that the input image contains white noise,
sophisticated parameter tuning,
and a significant amount of computational resources.

\section{Noise Estimation with RGB Correlations}
\subsection{Assumptions}
\label{sec:assumptions}
A model for a noisy RGB image with additive noise is given by
\begin{align}
	I_c = f_c + n_c \ \ \ \left(c \in \{R, G, B\}\right)
\end{align}
where $I_c$ is the observed noisy image, $f_c$ is the noise-free image (texture), and $n_c$ is noise.
We assume that textures are correlated between RGB channels 
and noise is uncorrelated with other signals in image patch $P_i$.
This is expressed as
\begin{align}
	\CovPi \left[ f_c, f_{c'} \right] &= S_{\mathit{cc'}} \\
	\En \left[ \CovPi \left[ f_c, n_{c'} \right] \right] &= 0 \\
	\En \left[ \CovPi \left[ n_c, n_{c'} \right] \right] &= 0 \ \ \ (c \neq c')
\end{align}
for $c, c' \in \{R, G, B\}$, 
where $\CovPi[\cdot, \cdot]$ is the covariance operator in patch $P_i$,
$S_{\mathit{cc'}}$ is the covariance between noise-free images in each channel, 
and $\En[\cdot]$ is the expected value with respect to noise $n$.
We also assume that each channel has the same noise level:
\begin{align}
	\En \left[ \VarPi \left[ n_c \right] \right] = \sigma^2 \ \ \ (c \in \{R, G, B\})
\end{align}
Here, $\VarPi [\cdot]$ is the variance operator in patch $P_i$, and $\sigma$ is the ground-truth noise level.

\subsection{Proposed Method}
\subsubsection{Noise-Level Estimation}
\label{sec:noise_level_estimation}
Let $\alpha_i$ and $\beta_i$ be the variables obtained
from randomly sampled image patch $P_i\,(i=1,2,\ldots,N_p)$ by calculating
\begin{align}
\alpha_i &= \frac{\VarPi[I_R] + \VarPi[I_G] + \VarPi[I_B]}{3} \\
\beta_i &= \VarPi \left[ \frac{I_R+I_G+I_B}{3} \right] 
\end{align}
In other words, 
variable $\alpha_i$ is the mean of the channelwise-variance of pixel values (mean of variance)
and variable $\beta_i$ is the variance of pixel values in the mean image (variance of mean).
The following equation and inequality hold for $\alpha_i$ and $\beta_i$:
\begin{align}
	\En \left[ \alpha_i \right] &= \frac{S_R^2 + S_G^2 + S_B^2}{3} + \sigma^2 \label{eq:alpha} \\
	\En \left[ \beta_i \right] &\leq \frac{S_R^2 + S_G^2 + S_B^2}{3} + \frac{1}{3} \sigma^2 \label{eq:beta}
\end{align}
where $S_c^2\ (c \in \{R, G, B\})$ is the variance of the noise-free image $f_c$ in patch $P_i$
(details are shown in the appendix).
Both sides of Inequality \ref{eq:beta} are equal if condition $C$ is satisfied:
\begin{align}
	\mbox{$C$: $f_c - f_{c'}\ (c\neq c') \ \ $is constant in each channel of patch $P_i$}
\end{align}
In other words, condition $C$ is equivalent to the following condition:
pixel values of a difference image between two channels are constant in patch $P_i$.
By substituting Equation \ref{eq:alpha} into Inequality \ref{eq:beta}, we have
\begin{align}
	\label{eq:sigma_upper_bound}
	\sigma^2 \leq  \En \left[ \frac{3}{2} (\alpha_i - \beta_i) \right]
\end{align}
with equality if condition $C$ is satisfied.

Inequality \ref{eq:sigma_upper_bound} shows that the right side is a good approximation
of the ground-truth noise level if condition $C$ is satisfied.
Therefore, noise-level estimate $\tilde{\sigma}_i^2$ for each patch $P_i$ is given by
\begin{align}
	\tilde{\sigma}_{i}^{2} = \frac{3}{2} (\alpha_i - \beta_i)
\end{align}
Now, we have the noise-level estimate of the entire image $\tilde{\sigma}^2$ as the weighted mean of $\tilde{\sigma}_{i}^{2}$:
\begin{align}
	\label{eq:weighted_mean}
    \tilde{\sigma}^2 = 
    \left( \sum_{i=1}^{N_p} w_i \tilde{\sigma}_{i}^{2} \right) \Bigg/
    \left( \sum_{i=1}^{N_p} w_i \right)
\end{align}
where $w_i$ is a weight whose value depends on to what extent condition $C$ is satisfied,
and $N_p$ is the number of sampled patches.
The details of weight determination are explained in the next section.

\subsubsection{Weight Determination}
\label{sec:weight_determination}
The accuracy of noise-level estimate $\tilde{\sigma}_i^2$ depends on to what extent condition $C$ is satisfied,
so we define loss $L_i$ for each patch as
\begin{align}
	L_i = \frac{\VarPi[f_R - f_G] + \VarPi[f_G - f_B] + \VarPi[f_B - f_R]}{3}
\end{align}
However, the exact value of loss $L_i$ cannot be obtained
since noise-free image $f_c\,(c \in \{R, G, B\})$ is unknown.
Therefore, we approximate loss $L_i$ by
\begin{align}
	\tilde{L}_i = \frac{\VarPi[I'_R - I'_G] + \VarPi[I'_G - I'_B] + \VarPi[I'_B - I'_R]}{3}
\end{align}
where $I'_c$ is an image blurred with a Gaussian filter with the standard deviation of $\sigma_{\mathrm{blur}}$.
This results in better noise-estimation performance than simply using the original image $I_c$
because blurred image $I'_c$ approximates noise-free image $f_c$ by removing the noise from original image $I_c$.
Note that the blurring does not affect the correlations between RGB channels
since the filter is independently applied to each channel with the same blur strength.

Weight $w_i$ should be large when the loss is small and vice versa, 
so we define weight $w_i$ as
\begin{align}
	w_i = \exp \left(
	-\gamma \frac{\tilde{L}_i}{\sum_{j=1}^{N_p} \tilde{L}_j / N_p}
	\right)
\end{align}
where $\sum_j \tilde{L}_j / N_p$ is the normalization factor of $\tilde{L}_i$,
and $\gamma$ is a parameter that determines how strongly patches with high losses are filtered out
(we also manually exclude patches 
if they contain an overexposed or underexposed area
because noise levels in such areas are considered smaller than the true noise level).
In Section \ref{sec:gamma_patch_size}, we experimentally show how parameter $\gamma$ affects noise-estimation performance.

\subsection{Extension for RAW Images}
\label{sec:extension_raw}
\begin{figure}[t]
    \begin{minipage}{0.49\hsize}
        \centering
        \includegraphics[width=0.6\linewidth]{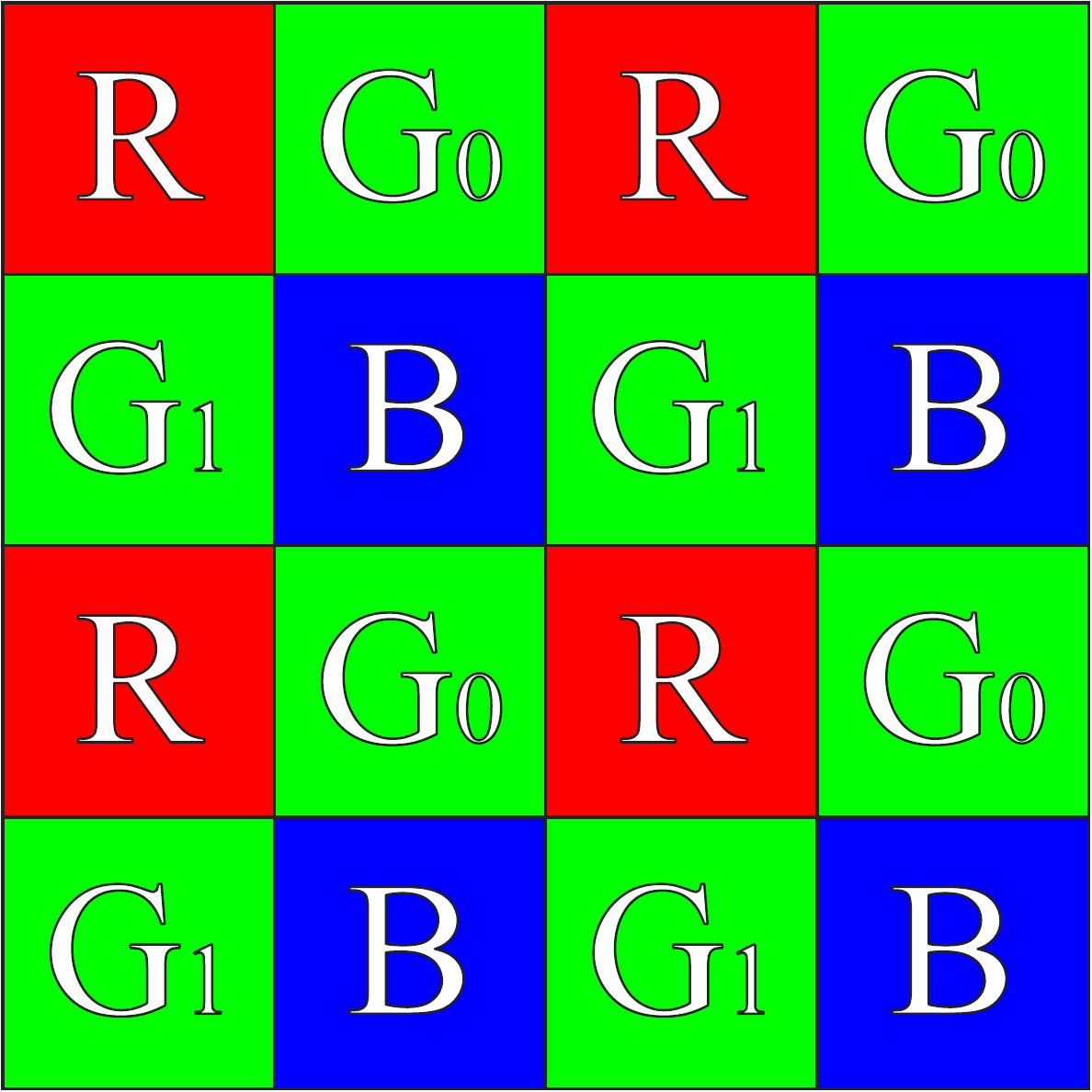}
    \end{minipage}
    \begin{minipage}{0.49\hsize}
        \centering
        \includegraphics[width=0.6\linewidth]{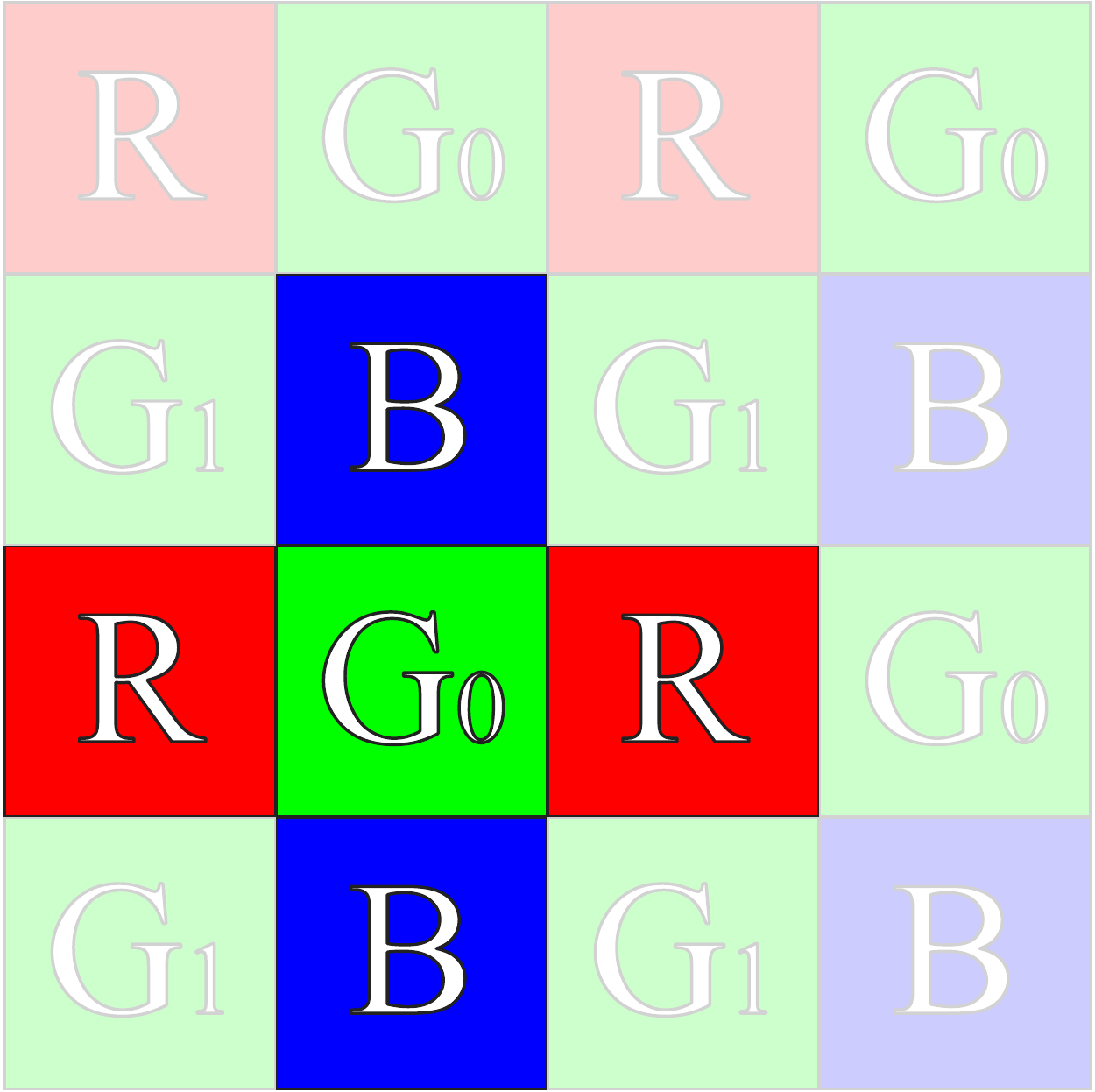}
    \end{minipage}
    \vspace{1.5mm}
    \caption{
        Left: Bayer pattern of RAW image. 
        Green cells are divided into two subgroups $G_0$ and $G_1$,
        so numbers of cells $R$, $G_0$, $G_1$ and $B$ are same.
        Right: We calculate red and blue components at each cell $G_0$  
        by taking average of neighbor cells. 
        In same way, red and blue components are calculated at each cell $G_1$.
    }
    \label{fig:bayer_pattern}
\end{figure}
In this section, we discuss extending our method for RAW images.
A RAW image contains unprocessed sensor outputs in the Bayer pattern, as shown in Figure \ref{fig:bayer_pattern}.
Note that the green cells are divided into two subgroups $G_0$ and $G_1$ 
so that each group ($R$, $G_0$, $G_1$, and $B$) has the same number of cells.
We interpolate the red and blue components at each green cell 
by averaging neighbor cells, as shown in Figure \ref{fig:bayer_pattern}.

Now, we obtain two sub-images $I^{(0)}$ and $I^{(1)}$ by extracting the RGB components
from subgroups $G_0$ and $G_1$, respectively.
Note that these sub-images are half the size of the original RAW image
since the sampling is carried out with the stride of two.
By concatenating patches $P_i^{(0)}$ and $P_i^{(1)}$ 
sampled in the same area from sub-images $I^{(0)}$ and $I^{(1)}$,
we obtain patch $P_i$, which can be treated in the same manner as that discussed in Section \ref{sec:noise_level_estimation}.
Note that the noise variance of the red and blue components in the sub-images 
is $\sigma^2 / 2$ since they are obtained by averaging two pixels of the original RAW image.
Therefore, Equation \ref{eq:alpha} and Inequality \ref{eq:beta} should be as follows:
\begin{align}
	\En \left[ \alpha_i \right] &= \frac{S_R^2 + S_G^2 + S_B^2}{3} + \frac{2}{3} \sigma^2 \label{eq:alpha_raw} \\
	\En \left[ \beta_i \right] &\leq \frac{S_R^2 + S_G^2 + S_B^2}{3} + \frac{2}{9} \sigma^2  \label{eq:beta_raw}
\end{align}
By substituting Equation \ref{eq:alpha_raw} into Inequality \ref{eq:beta_raw}, 
the noise-level estimate of each patch for RAW images is given as
\begin{align}
	\tilde{\sigma}_{i}^{2} = \frac{9}{4} (\alpha_i - \beta_i)
\end{align}

\section{Experiments}
In this section, we first discuss noise-estimation performance of the proposed method
for images artificially degraded with Gaussian noise.
Next, we analyze the relationship between parameter $\gamma$ and patch size,
which are both import parameters in the proposed method, 
to show that the noise-estimation performance of our method increases 
by using the weighted mean of the noise estimates rather than using the unweighted mean in Equation \ref{eq:weighted_mean}.
Then we evaluate the noise-estimation performance of our method for natural noisy images
and analyze the noise correlations in these images, which affects the noise-estimation performance of our method.
Finally, we compare the noise-estimation performance of our method to those of conventional methods.

In our experiments, we used JPEG and RAW images taken with a digital camera (Sony ILCD-7S).
We also generated lossless PNG images from the RAW images using ImageMagick \cite{image_magick}
to exclude the effects of JPEG compression.
For simplicity, we normalized the images by dividing them by 255 
so that all pixel values are within the range of $[0, 1]$.

\subsection{Evaluation}
\label{sec:evaluation}

\subsubsection{Artificially Degraded Images}
\label{sec:artificially_degraded_images_exp}
\begin{figure}[t]
    \centering
    \includegraphics[width=0.9\linewidth]{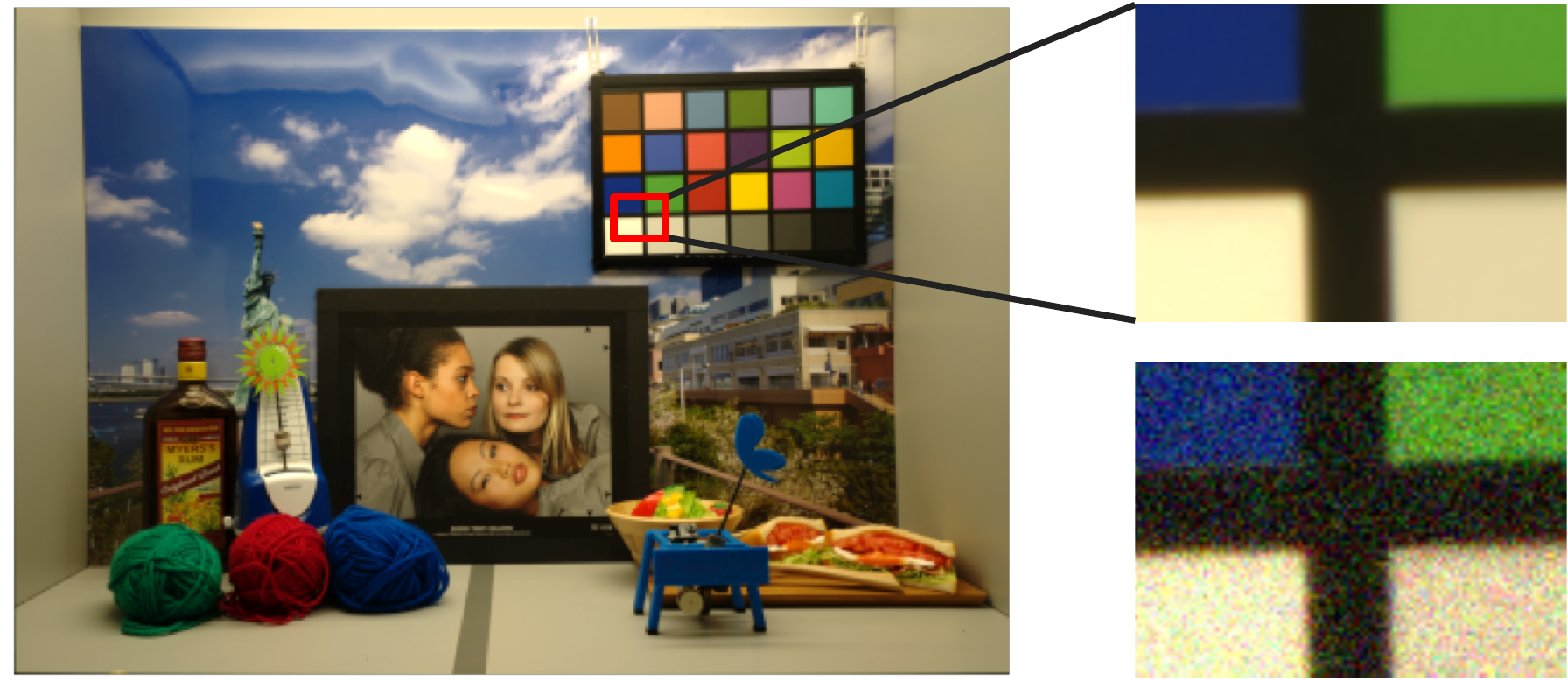}
    \caption{
        Left: Noise-free image obtained by taking average of 10 clear images.
        Upper right: Close-up of noise-free image.
        Lower right: Close-up of synthetic noisy image with Gaussian noise (ground-truth noise level: $\sigma = 0.095$).
    }
    \label{fig:close_up}
\end{figure}

\begin{figure}[t]
    \centering
    \includegraphics[width=0.8\linewidth]{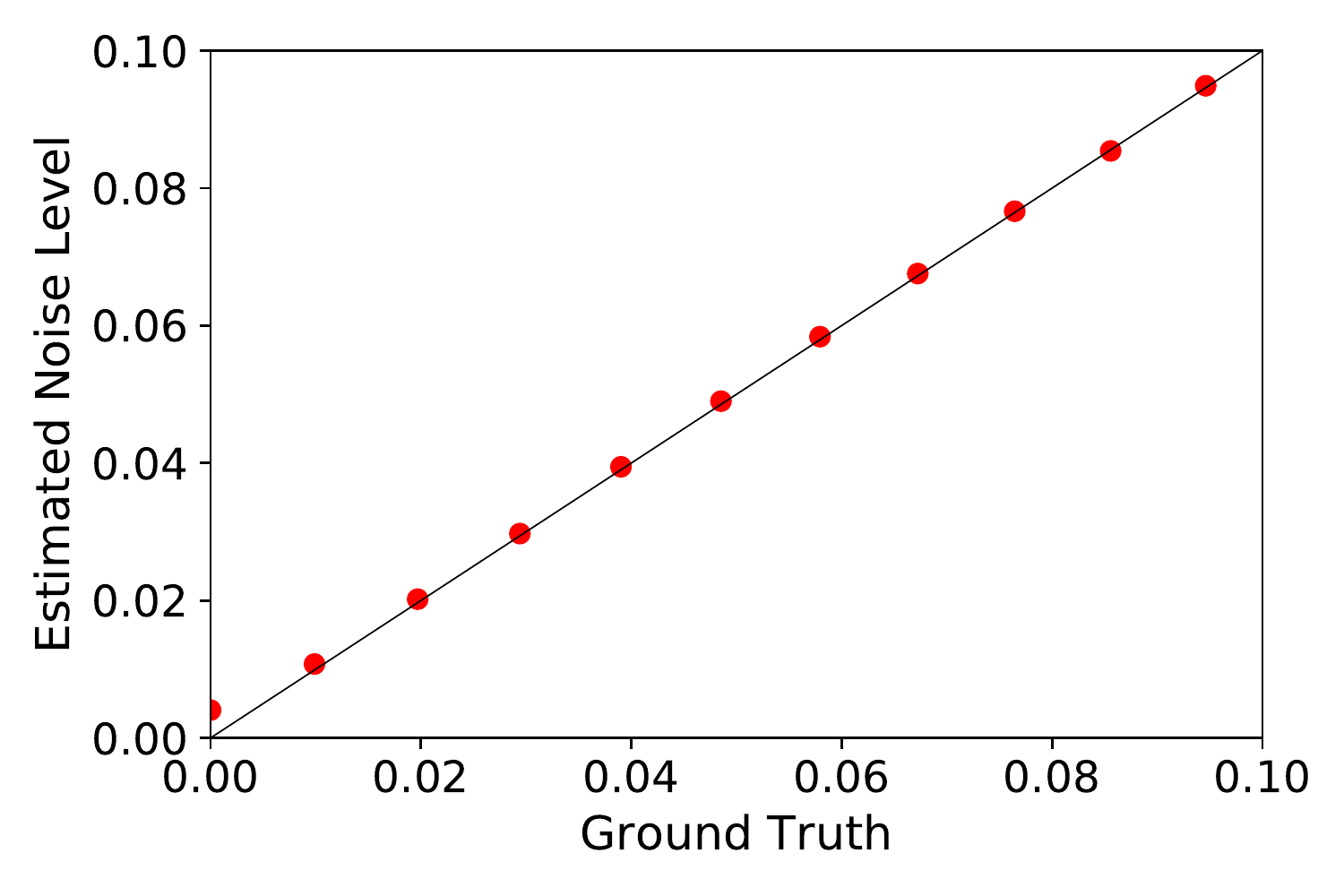}
    \caption{
        Relation between synthetic Gaussian-noise level
        and noise estimates with proposed method.
        Line $y=x$ in this figure shows ideal noise-estimation result.
    }
    \label{fig:artificially_degraded}
\end{figure}

We evaluated the noise-estimation performance of our method for synthetic noisy images
generated by adding Gaussian noise to a noise-free image (Figure \ref{fig:close_up}).
The noise-free image was generated by taking the average of 10 static images
taken under good photographic conditions (ISO: 50, exposure time: $2.5\,\mathrm{s}$).
We then added the Gaussian noise to the noise-free image 
and truncated the pixel values so that they stay in the range of $[0, 1]$.
We applied the proposed method to the synthetic noisy images 
and compared the noise estimates with the ground-truth noise levels.
The ground-truth noise level was obtained in the following manner:
first, we generated ten noisy images by adding Gaussian noise of the same noise level to the noise-free image;
second, we calculated the pixel value variances across the ten images at each pixel;
finally, we obtained the ground-truth noise level by taking the squared root of the mean of variances
calculated in the second step.
Note that the ground-truth noise level is slightly smaller than the standard deviation of the Gaussian noise
since the pixel values are truncated to stay in the range of $[0, 1]$.
The parameters were set as follows: 
$\gamma=2.0$, 
$k=5$,
$N_p=1000$, and
$\sigma_{\mathrm{blur}}=5.0$
where $\gamma$ is a parameter that determines how strongly patches with high losses are filtered out,
$k$ is the patch size,
$N_p$ is the number of sampled patches,
and $\sigma_{\mathrm{blur}}$ is the standard deviation of Gaussian filter used in weight determination.

Figure \ref{fig:artificially_degraded} shows that our method succeeded in accurately estimating the ground-truth noise level.
However, it slightly overestimated the noise level 
when the ground-truth noise level was very small ($\sigma \leq 0.01$).
This is considered to be the noise that could not be completely eliminated in the noise-free image.

\subsubsection{Relation Between Parameter $\gamma$ and Patch Size}
\label{sec:gamma_patch_size}
\begin{figure}[t]
    \centering
    \includegraphics[width=0.9\linewidth]{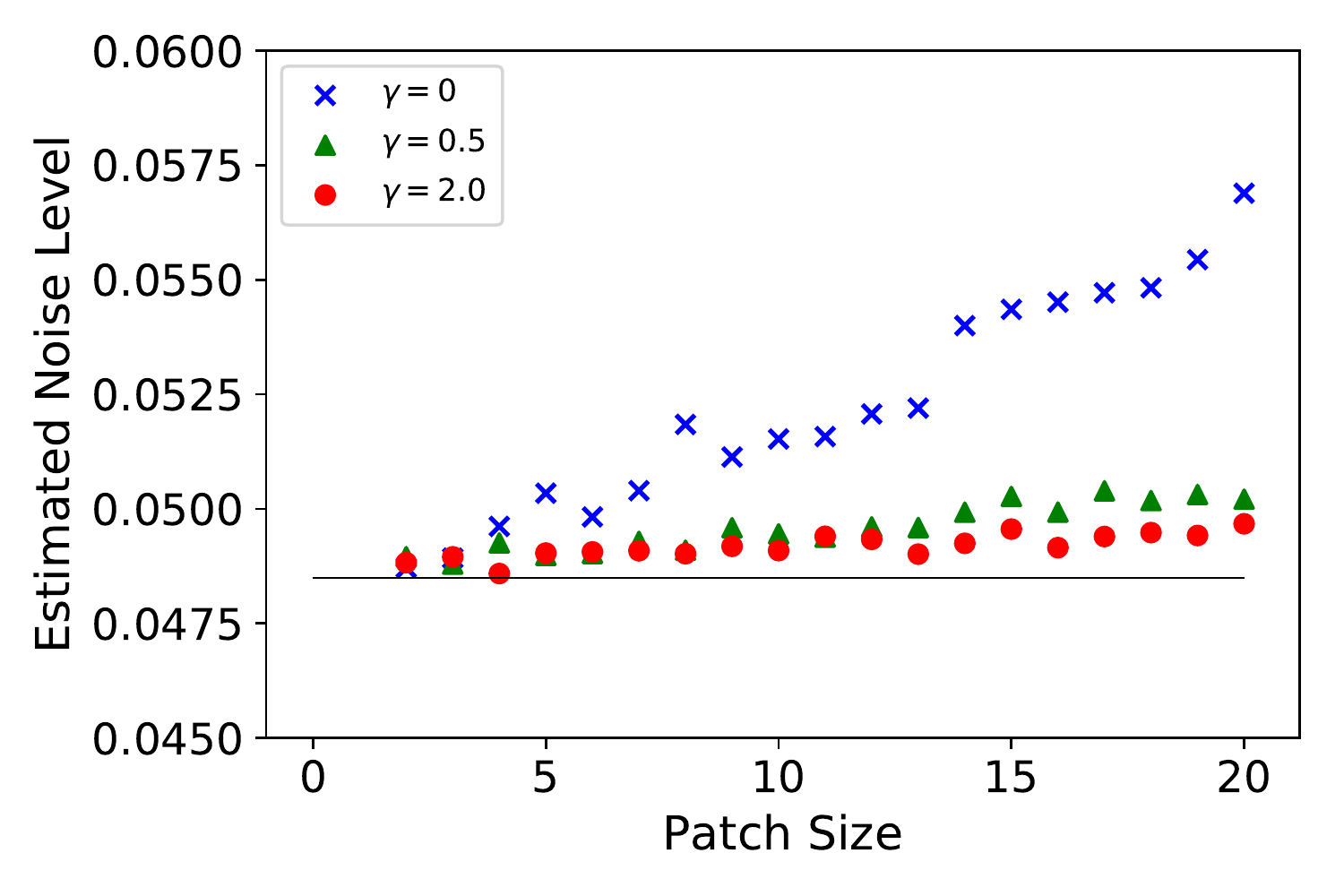}
    \caption{
        Relation between patch size and noise estimates	with different parameter $\gamma$ values.
        When parameter $\gamma$ is large enough, noise estimate is almost independent of patch size and closer to ground-truth noise level ($\sigma=0.0485$).
    }
    \label{fig:gamma_patch_size}
\end{figure}
In this section, we show that the noise estimate is independent of patch size 
if we use the weighted mean of the noise estimates (Equation \ref{eq:weighted_mean}), 
while the noise estimate is affected by patch size if we simply take the unweighted mean.
We analyzed the relation between the accuracy of noise estimates and patch size
while changing parameter $\gamma$.
The ground-truth noise level ($\sigma = 0.0485$) was calculated in the same manner as that discussed in Section \ref{sec:artificially_degraded_images_exp}.
We set the parameters as follows:
$\gamma \in \{0, 0.5, 2.0 \}$,
$k \in \{2, 3, 4, \ldots, 20\}$,
$N_p = 1000$, and
$\sigma_{\mathrm{blur}} = 5.0$.
As mentioned above, parameter $\gamma$ determines how strongly patches with high losses are filtered out,
so Equation \ref{eq:weighted_mean} is equivalent to taking the unweighted mean
when parameter $\gamma$ is set to zero.

Figure \ref{fig:gamma_patch_size} shows that the noise estimate is closer to the ground truth and independent of the patch size 
when parameter $\gamma$ is large enough.
This means that the noise-estimation performance of our method improved using the weighted mean rather than unweighted mean.
Interestingly, the noise estimate increases as the patch size increases
if we use the unweighted mean (\ie $\gamma = 0$).
This can be explained as follows.
If the patch size is large, we are more likely to have patches that do not satisfy condition $C$,
and this results in a larger expectation value of noise estimates, as shown in Inequality \ref{eq:sigma_upper_bound}.
We avoid this problem by using the weighted mean to exclude these patches.
Note that parameter $\gamma$ should not be too large 
because there is a lack of noise-estimate samples from filtering out most of the noise estimates.

\subsubsection{Natural Images}
\label{sec:natural_images_exp}
\begin{figure}[t]
    \centering
    \includegraphics[width=0.8\linewidth]{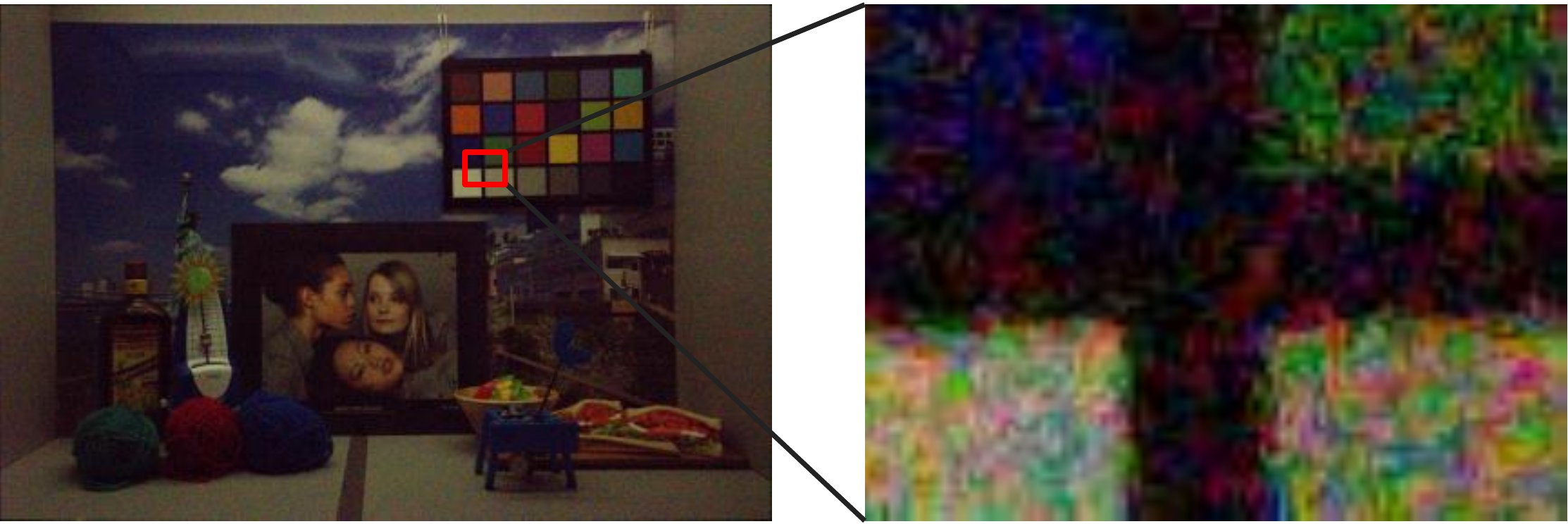}
    \includegraphics[width=0.8\linewidth]{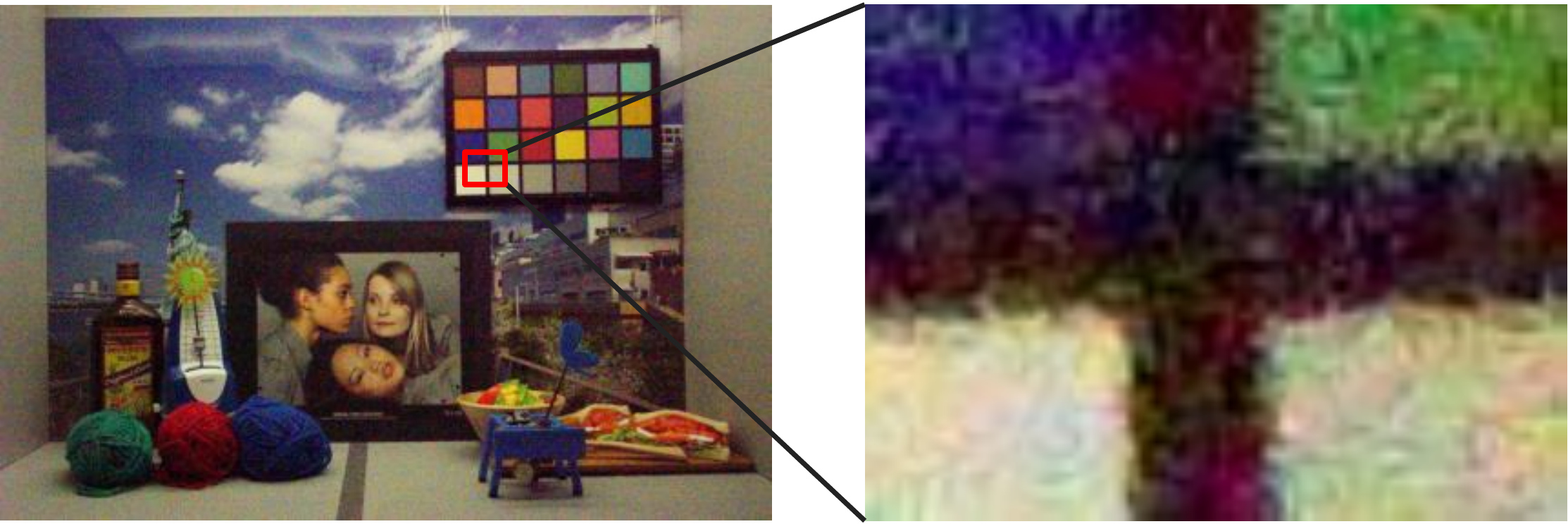}
    \caption{
        Sample natural noisy images used in experiments.
        Upper: Noisy PNG image and close-up.
        Lower: Noisy JPEG image and close-up.
    }
    \label{fig:natural_image_closeup}
\end{figure}
\begin{figure}[t]
    \centering
    \includegraphics[width=0.8\linewidth]{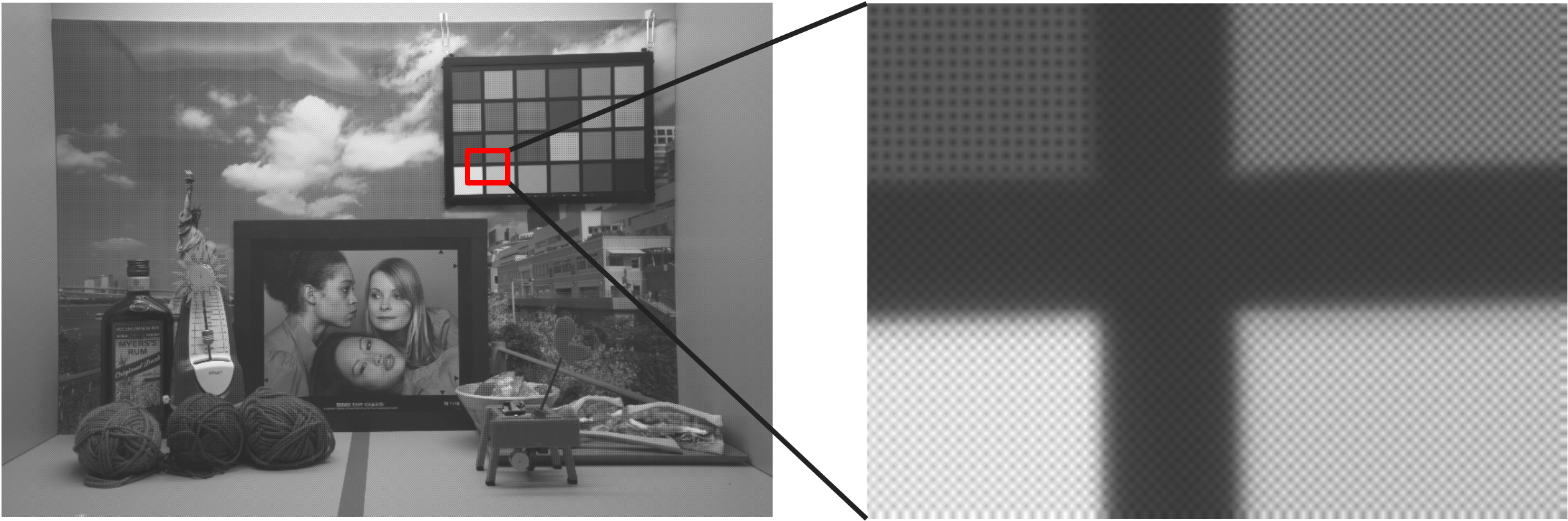}
    \includegraphics[width=0.8\linewidth]{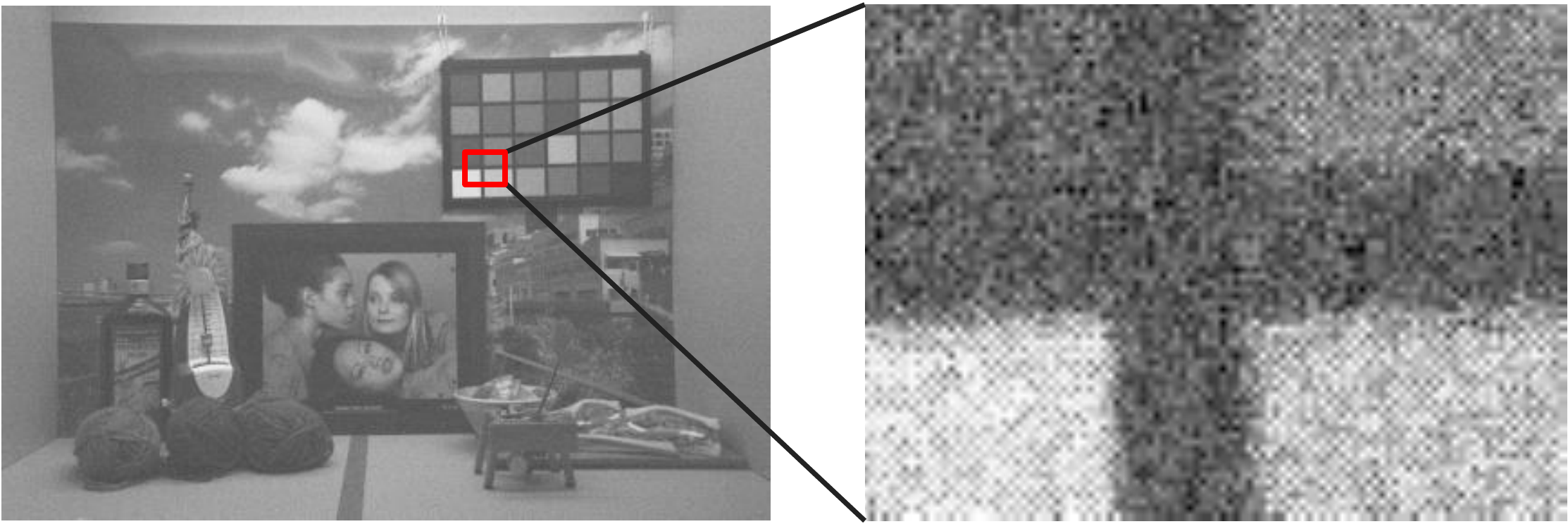}
    \caption{
        RAW images used in experiments.
        Note that RAW image can be visualized as gray-scale image.
        Upper: Clean RAW image and close-up.
        Bayer pattern can be observed, which is unique to RAW images.
        Lower: Noisy RAW image and close-up.
        Image is strongly degraded but Bayer pattern still can be observed
        (recommended to view this figure in electronic version
        to obtain clear view of Bayer pattern).
    }
    \label{fig:raw_image_closeup}
\end{figure}

\begin{figure}[t]
    \centering
    \includegraphics[width=0.85\linewidth]{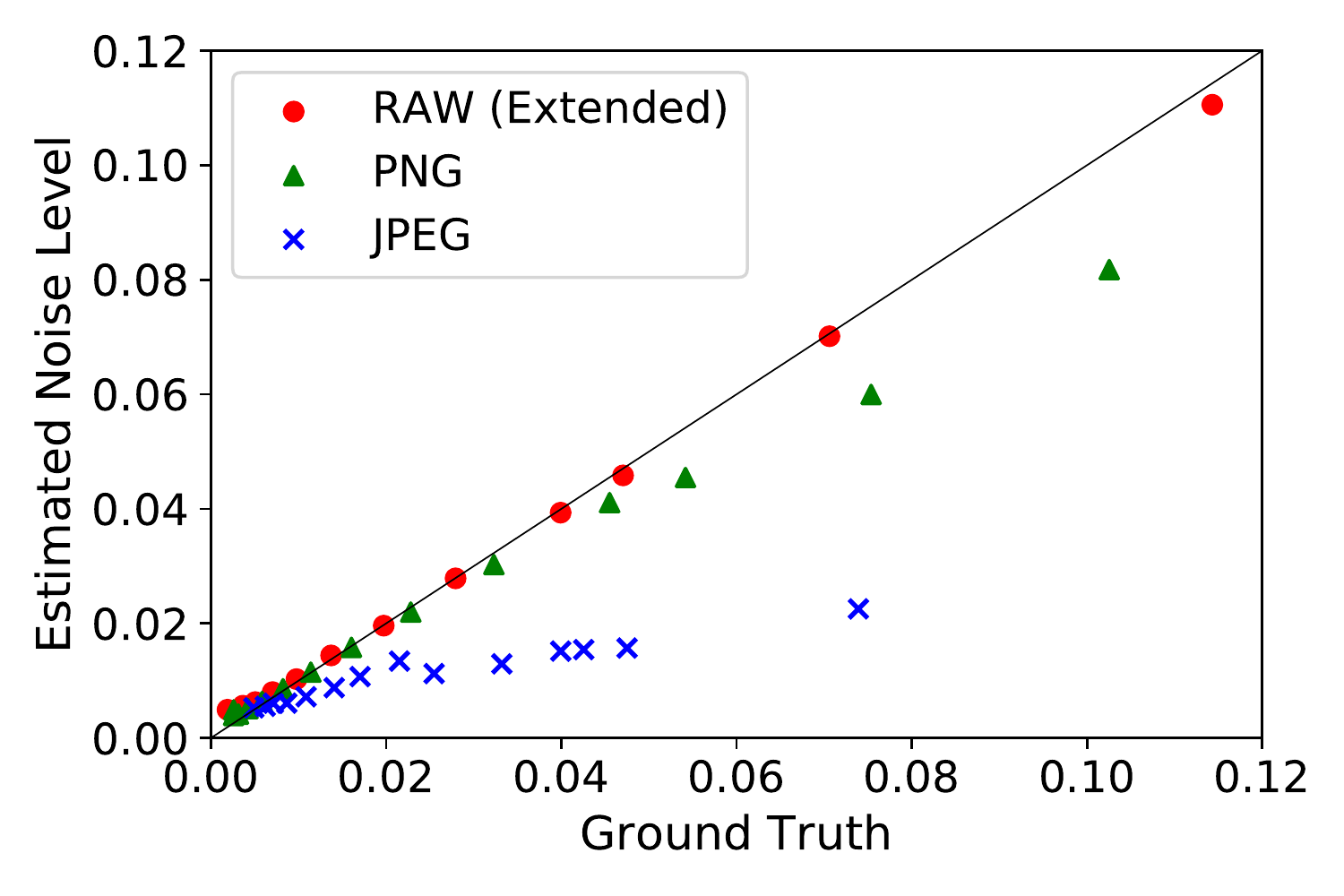}
    \caption{
        Results of noise estimation for natural images.
        Our extended method was applied to RAW images and original method was applied to PNG and JPEG images.
    }
    \label{fig:natural_images}
\end{figure}
We evaluated the noise-estimation performance of our method 
for natural noisy images (Figures \ref{fig:natural_image_closeup} and \ref{fig:raw_image_closeup}).
In this experiment, we used RAW, PNG, and JPEG images with various noise levels
under different photographic conditions
([ISO, exposure time] = 
[50, $1\,\mathrm{s}$] to [409600, $1/8000\,\mathrm{s}$]).
The RAW and JPEG images were directly obtained from the camera,
and the PNG images were generated from the RAW images.
We calculated the ground-truth noise level in the same manner as that discussed in Section \ref{sec:artificially_degraded_images_exp}
using 20 static images taken under the same photographic conditions.
The parameters were set as follows:
$\gamma = 2.0$,
$k =5$,
$N_p = 1000$, and
$\sigma_{\mathrm{blur}} = 5.0$.

Figure \ref{fig:natural_images} shows that 
our method estimated noise levels very accurately for the natural noisy RAW images
in a wide range of noise levels [0.0, 0.12].
However, the noise-estimation performance for the PNG and JPEG images was lower than
for the RAW images.
This is because of the noise correlations between RGB channels in natural noisy images.
We have so far assumed that noise is uncorrelated between RGB channels;
however, it has been shown that this assumption does not necessarily hold in natural images
because RGB channels are mixed up during in-camera processing 
and JPEG compressions \cite{gamut_mapping, noise_mixup}.
Note that our focus here is on the effect of JPEG compression on image noise,
rather than the noise called JPEG artifacts or {\it mosquito noise}.
Nam \etal \cite{noise_mixup} showed that 
in-camera processing and JPEG compression affect the noise characteristics, 
and the noise correlations caused by these processes cannot be ignored.
They also showed that the noise in each channel is almost uncorrelated in RAW images
(we analyze the noise correlations in natural images in Section \ref{sec:noise_correlations_exp}).
Therefore, the noise correlation is considered to be the main cause of the 
decrease in noise-estimate performance for natural PNG and JPEG images.
Although the noise-estimation performance of our method decreased in these images,
it is much better than those of the conventional methods, 
which are discussed in Section \ref{sec:comparison_natural}.

\subsection{Noise Correlations Between RGB Channels}
\label{sec:noise_correlations_exp}
\begin{table}[t]
	\caption{
		Noise correlation coefficients in natural images of different formats.
		RAW (0) and RAW (1) correspond to $I^{(0)}$ and $I^{(1)}$ in Section \ref{sec:extension_raw}, respectively.
	}
	\label{tab:noise_correlations}
	\centering
	\begin{tabular}{l|rrr}
		& $r_{RG}\ \ $ & $r_{GB}\ \ $ & $r_{GB}\ \ $ \\
		\hline 
		RAW (0) & 0.0031 & 0.0011 & 0.0025 \\
		RAW (1) & -0.0002 & 0.0041 & 0.0025 \\
		PNG & -0.0855 & -0.1414 & -0.0572 \\
		JPEG & 0.4852 & 0.5382 & 0.4755
	\end{tabular}
\end{table}

We analyzed the noise correlations in natural images.
We obtained a noise-free image by taking the average of 20 noisy static images 
(ISO: 409600, exposure time: $1/8000\,\mathrm{s}$)
and calculated the noise by subtracting the noise-free image from the original noisy image.
For RAW images, we used two sub-images $I^{(0)}$ and $I^{(1)}$ introduced in Section \ref{sec:extension_raw}
since RGB channels were not available in the original RAW images.
We calculated the noise correlations between RGB channels,
as shown in Table \ref{tab:noise_correlations}.
This shows that the noise correlations were high in the JPEG image ($|r| \approx 0.5$) 
while the RAW image had lower noise correlations ($|r| \approx 0.002$).
The PNG image had higher noise correlations than the RAW image ($|r| \approx 0.09$),
but was smaller than those of the JPEG image.
The cause of the high noise correlations in the JPEG and PNG images is considered to be 
the {\it developing} processes and JPEG compression, as Nam \etal \cite{noise_mixup} pointed out.

\subsection{Comparative Evaluation}
\label{sec:comparison}
We compared the noise-estimation performance of the proposed method
to those of conventional PCA-based methods proposed by Chen \etal \cite{improved_PCA} and Liu \etal \cite{PCA_1}.
We conducted two experiments for this comparison.
First, we compared the noise-estimation performance in synthetic noisy images degraded with Gaussian noise.
Next, we used natural noisy RAW, PNG, and JPEG images to compare its noise-estimation performance for practical use.

\subsubsection{Artificially Degraded Images}
\label{sec:comparison_artificial}
\begin{figure}[t]
    \centering
    \includegraphics[width=0.9\linewidth]{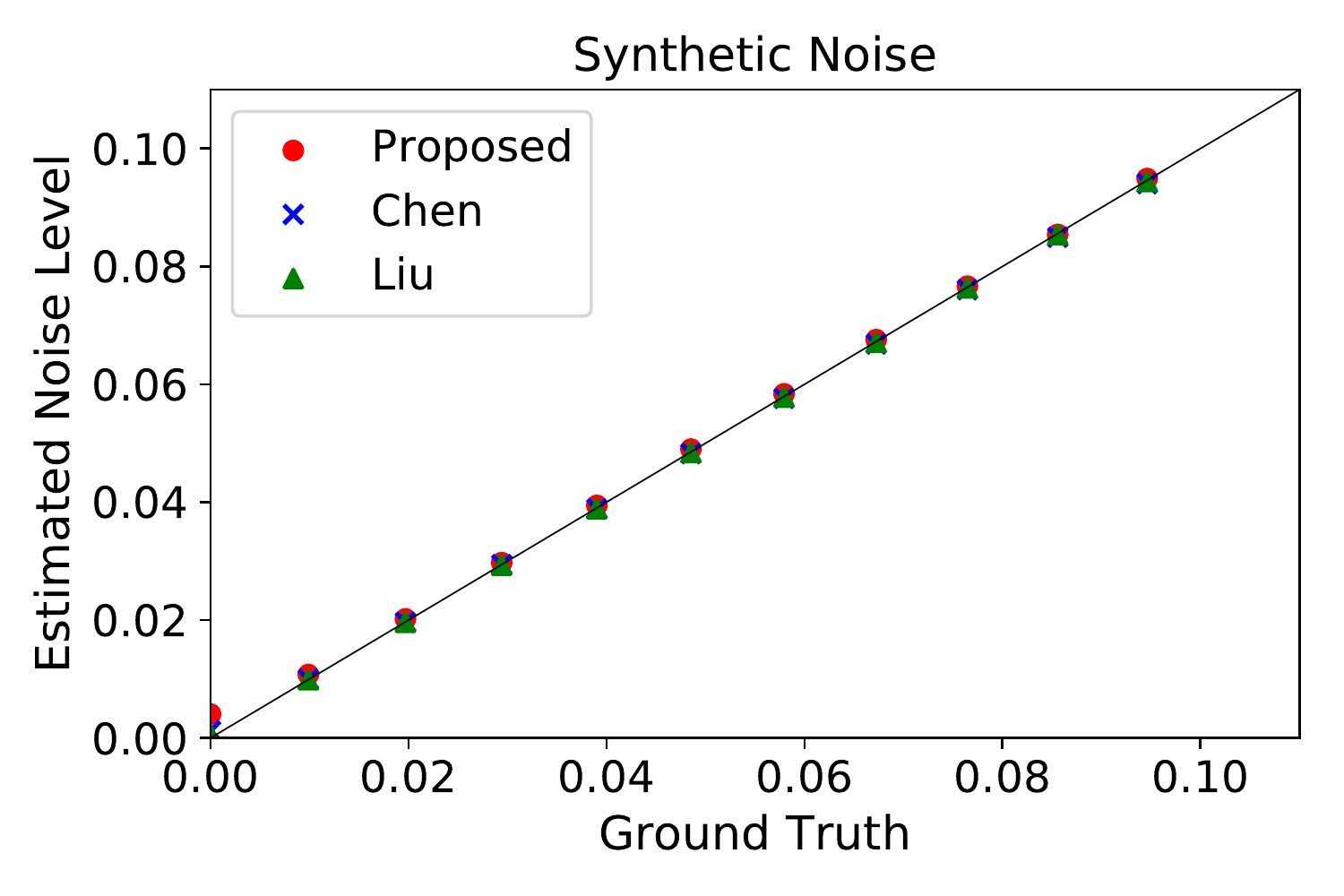}
    \caption{
        Comparison of noise-estimation performance for artificially degraded images.
        All methods achieved high noise-level-estimation performance.
    }
    \label{fig:compatison_synthetic}
\end{figure}
In the same manner as that discussed in Section \ref{sec:artificially_degraded_images_exp},
we evaluated the noise-estimation performance of our method for synthetic noisy images.
Figure \ref{fig:compatison_synthetic} shows that all methods succeeded in estimating the noise level very accurately.
Although the proposed method slightly overestimated the noise level compared to the conventional methods,
the estimation error was very small and can be ignored for practical use
(\eg when the ground-truth noise level $\sigma$ was 0.04852, 
the estimation error was 0.00047, -0.00028, and -0.00010
with the proposed method, Chen \etal's \cite{improved_PCA} and Liu \etal's \cite{PCA_1}, respectively.)

\subsubsection{Natural Images}
\label{sec:comparison_natural}

\begin{figure}[t]
    \centering
    \includegraphics[width=0.8\linewidth]{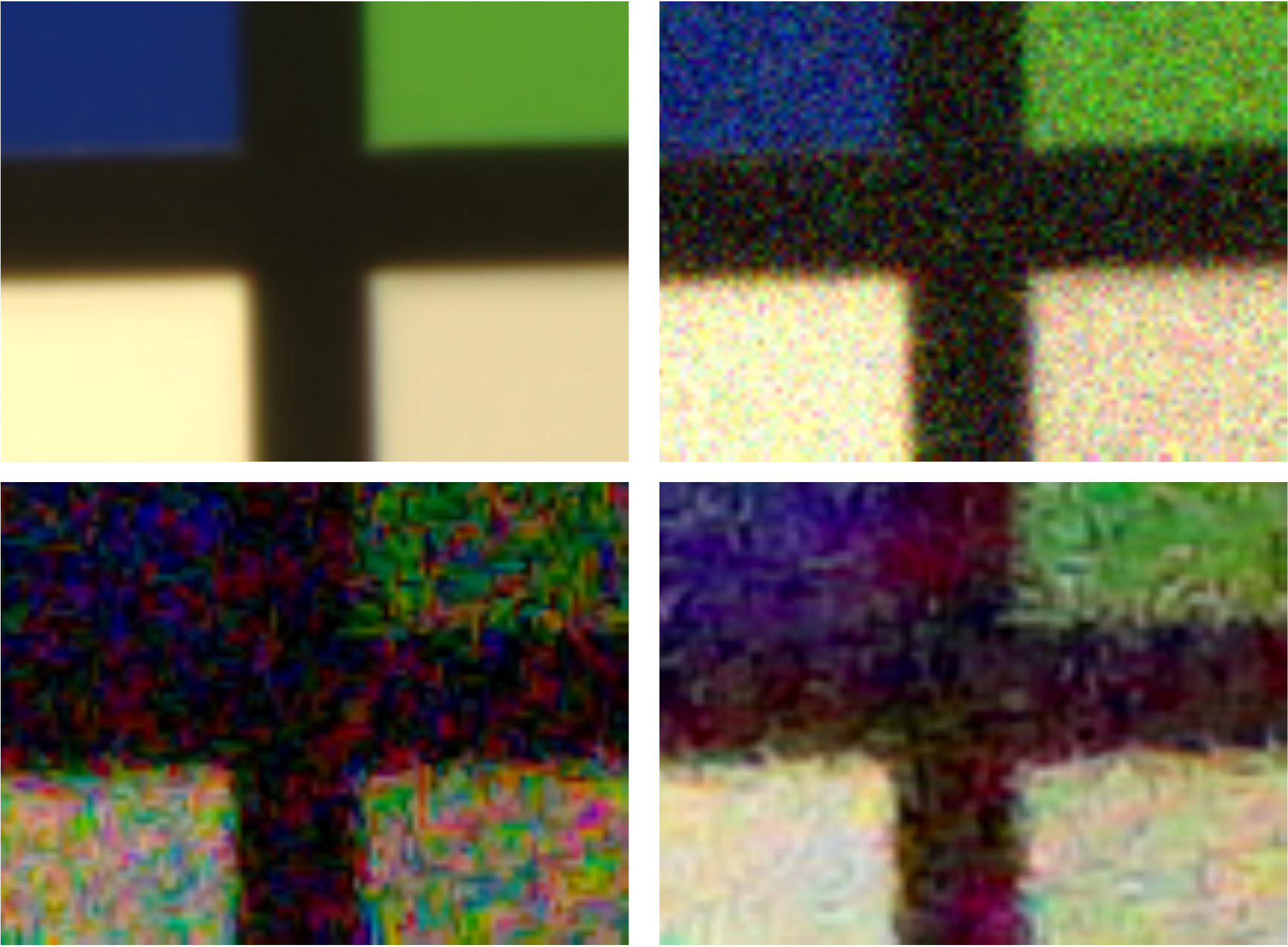}
    \caption{
        Close-up of images with same magnification ratio to compare noise patterns in detail.
        Upper left: Noise-free image.
        Upper right: Synthetic noisy image degraded with Gaussian white noise.
        Lower: Noisy PNG image (left) and noisy JPEG image (right).
        Recognizable spatial noise patterns can be observed in natural noisy images.
        This means that noise has spatial correlation in these images
        (recommended to view this figure in electronic version to obtain clear view of noise distribution).
    }
    \label{fig:noise_comparison}
\end{figure}
\begin{figure}[t]
    \centering
    \includegraphics[width=0.9\linewidth]{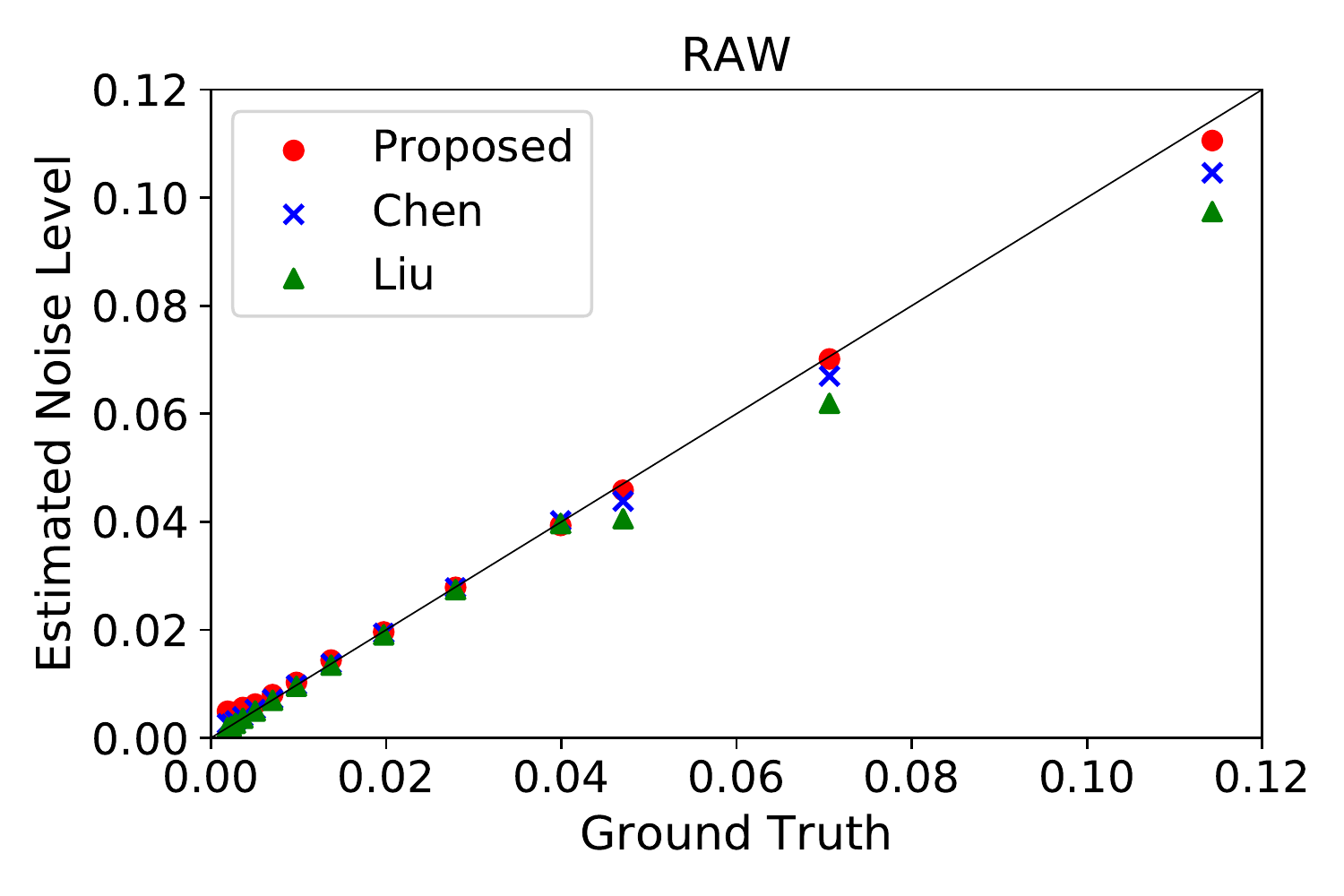}
    \includegraphics[width=0.9\linewidth]{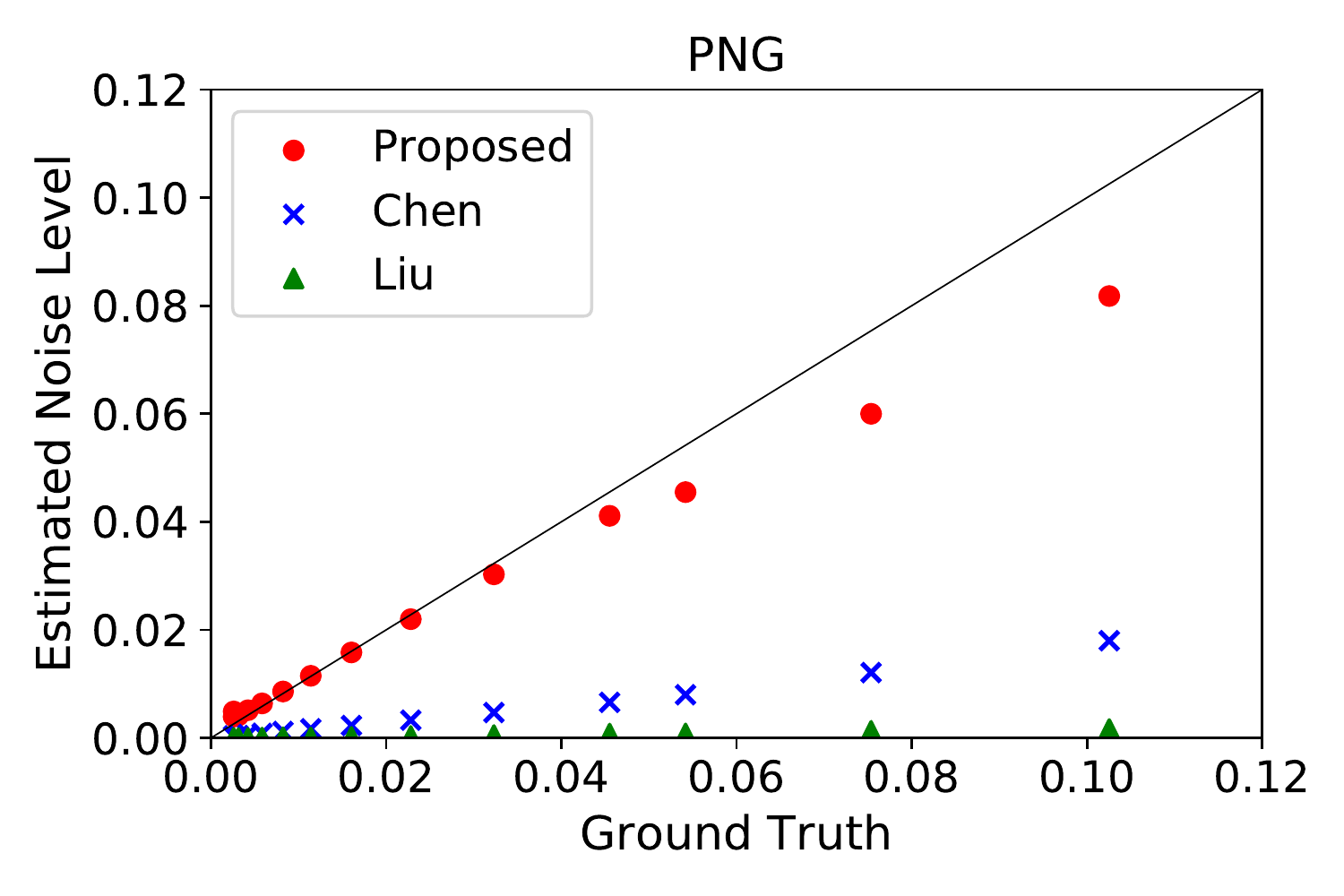}
    \includegraphics[width=0.9\linewidth]{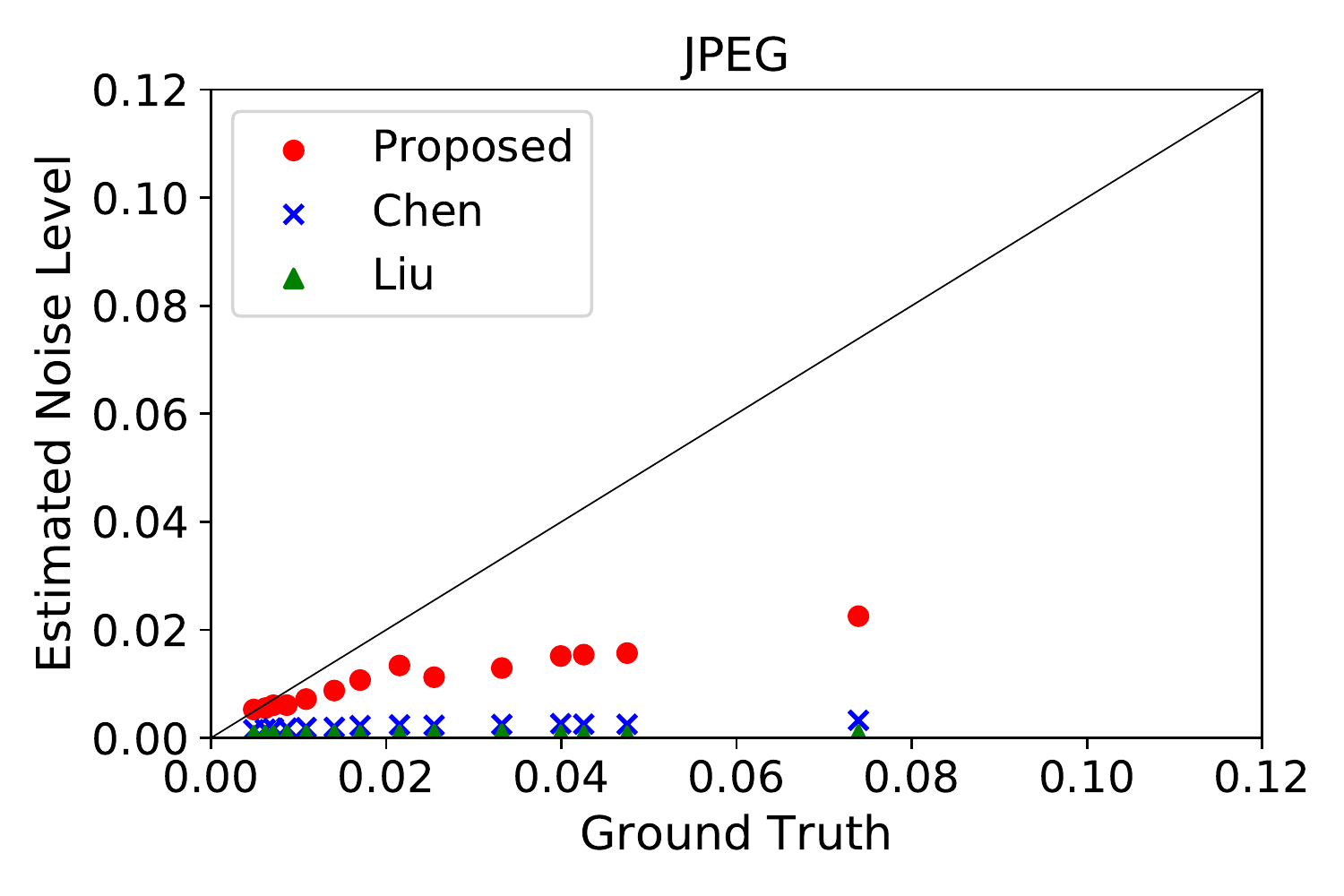}
    \caption{
        Comparison of noise-estimation performance in natural noisy images.
        Upper: Results from RAW images.
        Proposed method outperformed other methods,
        especially when ground-truth noise level was larger than 0.04.
        Middle: Results from PNG images.
        Proposed method better approximated ground-truth noise level
        while conventional methods failed.
        Lower: Results from JPEG images.
        All methods failed to estimate noise level,
        but noise estimates with proposed method were closer to ground truth
        compared to those of conventional methods.
    }
    \label{fig:comparison_natural_images}
\end{figure}

We compared the noise-estimation performance using natural noisy RAW, PNG, and JPEG images
in the same manner as that discussed in Section \ref{sec:natural_images_exp}.
The conventional methods cannot be directly applied to RAW images.
Therefore, we generated four gray-scale images by extracting pixel values
from each subgroup ($R$, $G_0$, $G_1$, and $B$) of the original RAW image,
and calculated the noise estimate by averaging the noise estimates obtained from the gray-scale images.
The results are shown in Figure \ref{fig:comparison_natural_images}.
Although our method tended to overestimate the noise level 
when the ground-truth noise level was very small ($\sigma \leq 0.005$),
it achieved higher accuracy than the conventional methods
in every image format,
which shows that our method is more suitable for practical use than conventional noise-estimation methods.
Interestingly, the conventional methods failed to estimate the noise level in the natural PNG and JPEG images.
These methods are based on the idea 
that the input image degrades with white noise.
However, natural noise has spatial correlations, as shown in Figure \ref{fig:noise_comparison},
so it cannot be regarded as white noise; hence, the low noise-estimation performance of the conventional methods.
On the other hand, the proposed method works well even if the input image is degraded with spatially correlated noise
since our method is based on the assumption that noise is uncorrelated in the channel direction rather than in the spatial direction.

\section{Conclusion}
We proposed a simple method 
of estimating noise level from a single color image
with prior knowledge that textures are correlated between RGB channels 
while noise is uncorrelated with other signals.
We also extended our method for RAW images
because they are useful and available in almost every digital camera.
We experimentally discussed the noise-estimation performance of the proposed method
for images degraded by synthetic Gaussian noise.
We also applied the proposed method to natural RAW, PNG, and JPEG images,
and it achieved higher noise-estimation performance than conventional noise-estimation methods.

Future work includes statistically analyzing the relationship between loss $\tilde{L}_i$ and noise estimates $\tilde{\sigma}_i$.
Weight $w_i$ is heuristically determined in the proposed method,
and it is not theoretically guaranteed that good patches are effectively selected with this weight.
Therefore, noise-estimation performance of our method can be further improved 
by statistically analyzing these variables.

\section*{Appendix: \\ Detailed Explanation of Section \ref{sec:noise_level_estimation}}
Let us define color channel set $U$ as $\{R, G, B\}$.
Variable $\alpha_i$ defined in Section \ref{sec:noise_level_estimation} is deformed as follows:
\begin{align}
\alpha_i 
&= \frac{1}{3} \sum_{c \in U} \VarPi[I_c] \\
&= \frac{1}{3} \sum_{c \in U} \VarPi[f_c + n_c] \\
&= \frac{1}{3} \sum_{c \in U} \left( \VarPi[f_c] + \VarPi[n_c] + 2\,\CovPi[f_c, n_c] \right)
\end{align}
By using the assumptions in Section \ref{sec:assumptions}, the following formula is obtained:
\begin{align}
\En \left[ \alpha_i \right] 
&= \frac{1}{3} \sum_{c \in U} \left( S_c^2 + \sigma^2 + 0 \right) \\
&= \frac{1}{3} \left(S_R^2 + S_G^2 + S_B^2 \right) + \sigma^2 \label{eq:appendix_e_alpha}
\end{align}
In the same manner, variable $\beta_i$ is deformed as follows:
\begin{align}
\beta_i 
&= \VarPi \left[ \frac{1}{3} \sum_{c \in U} I_c \right] \\
&= \frac{1}{9} \ \VarPi \left[ \sum_{c \in U} (f_c + n_c) \right] \\
&= \frac{1}{9}\ \sum_{c \in U} \left( \VarPi \left[ f_c \right] + \VarPi \left[ n_c \right] \right) \nonumber \\
&+ \frac{1}{9} \sum_{c,c' \in U \atop c \neq c'} \left( \CovPi\! \left[ f_c, f_{c'} \right] + \CovPi\! \left[ n_c, n_{c'} \right] \right) \nonumber \\
&+ \frac{2}{9} \sum_{c,c' \in U} \CovPi\! \left[ f_c, n_{c'} \right]
\end{align}
By using the assumptions in Section \ref{sec:assumptions}, we obtain
\begin{align}
\En \left[ \beta_i \right] 
&= \frac{1}{9}\ \sum_{c \in U} \left( S_c^2 + \sigma^2 \right)  \nonumber \\
&+ \frac{1}{9} \sum_{c,c' \in U \atop c \neq c'} \left( S_{cc'} + 0 \right) \nonumber \\
&+ \frac{2}{9} \sum_{c,c' \in U} 0 \\
&= \frac{1}{9} \left( S_R^2 + S_G^2 + S_B^2 + 2S_{\mathit{RG}} + 2S_{\mathit{GB}} + 2S_{\mathit{BR}} \right) \nonumber \\
&+ \frac{1}{3} \sigma^2
\end{align}
We apply Cauchy-Schwarz inequality $S_{\mathit{cc'}} \leq S_c S_{c'}$ and obtain
\begin{align}
\En \left[ \beta_i \right] 
&\leq \frac{1}{9} \left( S_R^2 + S_G^2 + S_B^2 + 2S_R S_G + 2S_G S_B + 2S_B S_R \right) \nonumber \\
&+ \frac{1}{3} \sigma^2 \label{eq:appendix_e_beta}
\end{align}
with equality if correlation coefficients $r_{\mathit{RG}}$, $r_{\mathit{GB}}$, and $r_{\mathit{BR}}$ are equal to 1.
We take the difference between Equation \ref{eq:appendix_e_alpha} and Inequality \ref{eq:appendix_e_beta} as follows:
\begin{align}
\En \left[ \alpha_i - \beta_i \right] 
&\geq \frac{2}{9} \left( S_R^2 + S_G^2 + S_B^2 - S_R S_G - S_G S_B - S_B S_R \right) \nonumber \\
&+ \frac{2}{3} \sigma^2 \\
&= \frac{1}{9} \Big( (S_R - S_G)^2 + (S_G - S_B)^2 + (S_B - S_R)^2 \Big) \nonumber \\
&+ \frac{2}{3} \sigma^2 \\
&\geq \frac{2}{3} \sigma^2 \ \ \ \mbox{(equality holds when $S_R=S_G=S_B$)}
\end{align}
Now, we obtain the following inequality
\begin{align}
\sigma^2 \leq  \En \left[ \frac{3}{2} (\alpha_i - \beta_i) \right]
\end{align}
with equality if the following conditions are satisfied:
\begin{align}
\begin{cases}
\label{eq:condition}
r_{\mathit{RG}} = r_{\mathit{GB}} = r_{\mathit{BR}} = 1 \\
S_R=S_G=S_B
\end{cases}
\end{align}
Condition \ref{eq:condition} is equivalent to the following condition:
\begin{quote}
	Centered noise-free images $f_R$, $f_G$, and $f_B$\\
	are the same in patch $P_i$.
\end{quote}
This can also be expressed as
\begin{quote}
	$f_c - f_{c'}\ (c\neq c') \ \ $is constant in each channel of patch $P_i$.
\end{quote}
This is equivalent to condition $C$ introduced in Section \ref{sec:noise_level_estimation}.


{\small
\bibliographystyle{ieee}
\bibliography{egbib}
}

\end{document}